\documentclass[conference]{IEEEtran}
\IEEEoverridecommandlockouts
\usepackage{cite}
\usepackage{amsmath,amssymb,amsfonts}
\usepackage{algorithm}
\usepackage{algpseudocode}
\usepackage{graphicx}
\usepackage{subfig} 
\usepackage{caption}
\usepackage{balance}
\usepackage{listings}
\usepackage{comment}
\usepackage{textcomp}
\def\BibTeX{{\rm B\kern-.05em{\sc i\kern-.025em b}\kern-.08em
    T\kern-.1667em\lower.7ex\hbox{E}\kern-.125emX}}
\begin{document}
\title{Introducing Federated Learning into Internet of Things ecosystems -- preliminary considerations\\
\thanks{This work is part of ASSIST-IoT project that has received funding from the European Union's Horizon 2020 research and innovation program under grant agreement 957258. Work of Maria Ganzha and Anastasiya Danilenka is funded in part by the Centre for Priority Research Area Artificial Intelligence and Robotics of Warsaw University of Technology within the Excellence Initiative: Research University (IDUB) programme.}
}

\author{\IEEEauthorblockN{Karolina Bogacka}
\IEEEauthorblockA{\textit{Systems Research Institute}\\\textit{Polish Academy of Sciences} \\
Warsaw, Poland \\
0000-0002-7109-891X}
\and
\and
\IEEEauthorblockN{Katarzyna Wasielewska-Michniewska}
\IEEEauthorblockA{\textit{Systems Research Institute}\\\textit{Polish Academy of Sciences} \\
Warsaw, Poland \\
0000-0002-3763-2373}
\and
\IEEEauthorblockN{Marcin Paprzycki}
\IEEEauthorblockA{\textit{Systems Research Institute}\\\textit{Polish Academy of Sciences} \\
Warsaw, Poland \\
0000-0002-8069-2152}
\and
\IEEEauthorblockN{Maria Ganzha, Anastasiya Danilenka}
\IEEEauthorblockA{\textit{Faculty of Mathematics and Information Science} \\
\textit{Warsaw University of Technology}\\
Warsaw, Poland \\
0000-0001-7714-4844, 0000-0002-3080-0303}
\and
\IEEEauthorblockN{Lambis Tassakos}
\IEEEauthorblockA{\textit{TwoTronic Gmbh} \\
Meitingen, Germany \\
0000-0003-2511-9035}
\and
\IEEEauthorblockN{Eduardo Garro}
\IEEEauthorblockA{\textit{Prodevelop} \\
Valencia, Spain \\
0000-0002-8160-0125}
}

\maketitle

\begin{abstract}

Federated learning (FL) was proposed to facilitate the training of models in a distributed environment. It supports the protection of (local) data privacy and uses local resources for model training. Until now, the majority of research has been devoted to ``core issues'', such as adaptation of machine learning algorithms to FL, data privacy protection, or dealing with the effects of uneven data distribution between clients. This contribution is anchored in a practical use case, where FL is to be actually deployed within an Internet of Things ecosystem. Hence, somewhat different issues that need to be considered, beyond popular considerations found in the literature, are identified. Moreover, an architecture that enables the building of flexible, and adaptable, FL solutions is introduced.

\end{abstract}


\section{Introduction}

One of the critical (and practical) bottlenecks of the application of Machine Learning (ML) lies in the limited ability to collect, consistently label, and use large datasets. This is particularly the case for businesses that do not possess almost unlimited resources, as Google or Amazon do~\cite{survey}. Moreover, while existing data may be large and labeled, it may be ``split between stakeholders'', who do not want to and/or cannot share their datasets~\cite{data-collection}. For instance, this is the case for the medical data, which belongs to different hospitals/clinics. Moreover, there are ongoing controversies concerning the collection and storage of information~\cite{blake}. Many ML developments, e.g. in mobile applications, rely on the models being periodically (re/up)trained on sensitive private data (e.g., browsing history, or geo-positioning). Hosting such data in a centralized location, even in adherence to strict legislation, still poses serious security risks, as can be seen through repeated data leaks~\cite{9144394,privacy,blockchain2}.

It is also worth noting that the latest advancements in ML involve training very large models that require enormous computational resources~\cite{1313}. This not only increases the cost but also the carbon footprint~\cite{DBLP:journals/corr/abs-2102-07627}.

To overcome these, and other related, problems, Federated Learning (FL) has been proposed~\cite{DBLP:journals/corr/McMahanMRA16}. The name of the approach came from the use of a flexible \textit{federation} of collaborating (often heterogeneous, edge) devices, known as clients, ``synchronized'' and ``orchestrated'' by a ``central server''. In FL, (i) clients train copies of the global model, using local data, and (ii) send updates to the server, which (iii) aggregates them, and (iv) updates the shared model~\cite{623}, which (v) is sent back to the clients to continue the process, until a stopping criterion is met. Therefore, private data never leaves the clients~\cite{fSM}. While a lot of research is devoted to the FL process itself, it is mostly implemented and tested in a cloud. This means that important practical issues that, as we will argue, have to be resolved, are omitted~\cite{9220170,DBLP:journals/corr/abs-2009-13012}. 

One should immediately realize that one of the future areas of application of FL is the Internet of Things. Among others, this is the result of a general trend to replace cloud-centric solutions with edge-cloud continuum-based approaches~\cite{challenges}. This is happening because storing data, and providing resources, in the data center is not sustainable for large-scale complex deployments, where latency can negatively impact performance. Hence, computing has to take place near (at) the edge of the network, physically close to sensors and/or users~\cite{edgeF}. The resulting ecosystem represents the edge-cloud continuum and is the necessary direction for the evolution of Next-Generation Internet of Things deployments~\cite{DBLP:journals/corr/abs-2009-13012}. Here, among others, FL will deliver intelligence at the edge~\cite{623}. However, combining FL with IoT brings about its own issues: (i) heterogeneity of clients and networks can cause delays (latency variability), or the presence of ``stragglers'' (weaker/more busy clients); (ii) computing and/or storage resources on the (far) edge devices, as well as their battery life, tend to be very limited, which impedes the use of large models and poses restrictions on training time; and (iii) data used for the training can be highly redundant~\cite{9220170}.

As noted, core research on Federated Learning is focused on machine learning (ML) and its intricacies. This can be seen also when one considers state-of-the-art of FL frameworks. For example, though TensorFlow Federated Framework (TFF)~\cite{tff} offers a wide variety of stable ML models, it supports experimentation only in a simulated environment. In other words, TFF currently does \textit{not} enable use of actual ``edge devices''. Another widely known FL platform is FATE~\cite{fate}. Here, 6GB of RAM, and 100 GB of disk space, on the server as well as on the clients are expected. While this would work in a laboratory, it exceeds the capabilities of the majority of edge devices (at least of today). Among platforms, PaddleFL enables the implementation of decentralized architectures by default. However, due to the low number of current contributors, and the employment of PaddlePaddle, a lesser-known Deep Learning platform~\cite{fltools}, PaddleFL may not be an optimal choice for future work. Flower (A Friendly Federated Learning Framework~\cite{flower2}) can be run on a diverse range of environments and devices, including Android, iOS, Raspberry Pi, and Nvidia Jetson. It is also compatible with popular ML frameworks like PyTorch and Keras. Finally,  PySyft~\cite{pysyft} allows the use of clients on the edge, using \textit{pygrid}, which is a novel development. However, even the latest two platforms can be seen, primarily, as tools for studying the ``nature of FL'', rather than to be used to run FL in IoT ecosystems.

In this context, this work aims to (a) reflect on the nature of challenges that actual FL deployments in IoT have to address, (b) show how a reference architecture, proposed for Next-Generation IoT supports the deployment of Federated Learning, and (c) illustrate the flexibility of the proposed approach through its capability of setting systems with different FL topologies. 
Hence, the remaining parts of this work are organized as follows. In Section~\ref{usecase} a practical IoT-based scenario from ASSIST-IoT~\footnote{https://assist-iot.eu/} project is described. Since FL will be actually deployed and experimented with in this use case, it will be used to summarize key requirements for ``practical FL in IoT''. In Section~\ref{topologies} we summarize pertinent state-of-the-art. Following, in Section~\ref{architecture}, an architecture that fulfills the requirements of the use case and addresses issues materializing in IoT-based deployments is described. Next, in Section~\ref{testing}, the ways in which the proposed architecture can be adapted and used are outlined. Finally, in Section~\ref{conclusions}, a summary of contributions, and directions of future work are provided.

\section{Federated Learning use case in IoT deployment}
\label{usecase}

The foundation of this contribution is provided by ASSIST-IoT project. There, a sample use case, in which FL is to be applied, is a part of the car damage recognition pilot.

The main goal of this scenario is to provide a fast and accurate inspection of car exterior damage, with minimal data transfer from edge devices to the cloud. Here, the task of car damage detection can be separated into three steps: (i) efficiently separating the vehicle from the background, (ii) vehicle part segmentation, and (iii) automatic defect detection. The results are to be used to support expert-delivered-evaluation, and to facilitate decisions involving insurance claims, as well as car return or leasing services. 

\begin{figure}[htbp]
\centering
\hfill
\subfloat{
\includegraphics[width=0.9\linewidth]{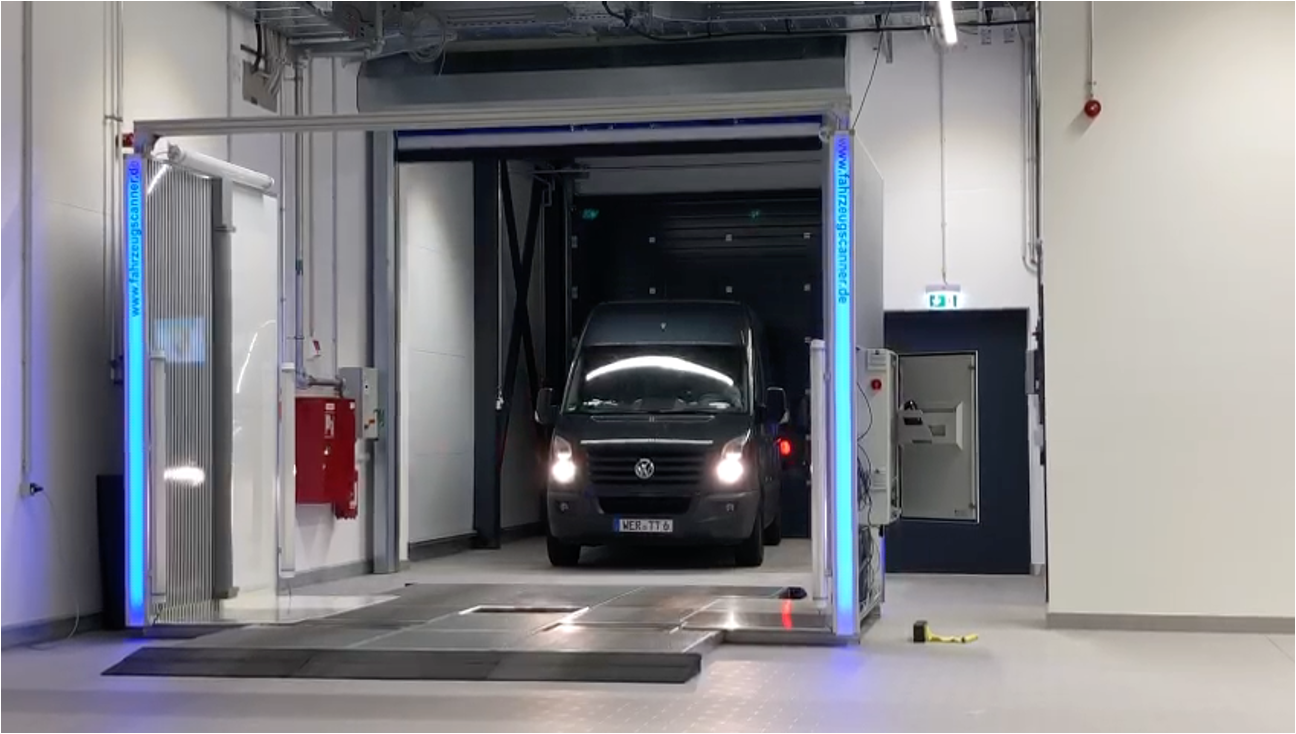}
}
\caption{Car damage recognition - scanner gate}
\label{Usecase_Figure}
\end{figure}

As it can be seen in Fig.~\ref{Usecase_Figure}, the functional pipeline involves multiple professional scanners, equipped with high-quality cameras, based on the TwoTronic solution~\footnote{https://www.fahrzeugscanner.de/}. A high volume (more than 200) of scanned vehicles per day is expected. TwoTronic scanners, and ``attached'' medium-class computers, will serve as FL clients. The FL server will be located in an external data center in Nürnberg, Germany.

Deploying a FL system is a complex task, depending not only on the availability of FL libraries and algorithms but also characteristics and limitations of a distributed system. Importantly, to be able to practically apply FL solution in this real-life use case, additional issues that are rarely addressed in literature, such as: (a) sudden user dropout, (b) weak network connection with potential interruptions, (c) geographical constraints (leading to unequal groups of clients), (d) data distribution (local distribution on the client differing from global distribution, with no additional public information that would enable problem mitigation through client grouping), and (e) system limitations, notably available RAM and number of cores, need to be considered. Lastly, (f) in environments with heterogeneous devices, interoperability may also become an issue. 

It is worth noting that, due to (geographical) distances between scanners and the FL server, located in Nürnberg, as well as the high speed and accuracy of prediction, necessary for this scenario, examining different Fl topologies may be in order. First, divergence from a centralized (client-server) schema to a decentralized one could protect the system from having a single point of failure. This could increase its reliability and resilience. Second, the introduction of additional aggregating clients into a centralized system, would mean that more information about an interrupted training is being preserved. Moreover, training could continue within lower levels of aggregating hierarchy. Additionally, a decrease in direct communication between scanners and the central server could mean faster training. The employment of non-standard topologies may also reduce sensitivity of the training process to interruptions and sudden client dropouts; by introducing additional communication channels. 

\section{Work related to Federated Learning topology}
\label{topologies}

Taking into account potential importance of FL topology, let us summarize related state-of-the-art. Currently, the effect of topology between clients on FL systems is not fully understood, but hard to deny~\cite{DBLP:4}. For sure, there is no ``best topology'', but rather it needs to be selected to match the characteristics of a specific use case. It has been observed that the centralized approach may not be appropriate, due to significant communication overhead and a single point of failure~\cite{efficient}. On the other hand, fully decentralized topologies can involve a significant cost of communication not related to client-to-server one~\cite{bacombo}. It is worth mentioning that some works combine these approaches to improve convergence and scalability, for example by combining decentralized groups with a centralized update schema~\cite{DBLP:journals/corr/abs-2012-03214}.

Some approaches experimented with star and ring architectures and their combination. The reason was to avoid the communication bottleneck of the former while gaining improved scalability and accuracy of the latter~\cite{Duan}. There, a star architecture with ring-based groups, supported by a self-balancing framework designed to mitigate the problem of a skewed global distribution, was evaluated.

Work presented in~\cite{DBLP:2} uses a ring architecture with star-based groups, in a realistic use case with non-IID data with periodic variance. Overall, while a linear speedup with respect to the number of clients is reported, the need for periodic variance is a limiting factor.

Ring-based groups, without global communication, while further elaborating on the periodically variational distribution of the data samples, treated by semi-cyclic Stochastic Gradient Descent (SGD) is discussed in~\cite{DBLP:3}. Here, it is observed that the use of ring-based groups may lead to slower training due to the higher number of rounds the process has to undergo for the model to gather information from all the nodes belonging to the group when compared with star-based groups.

Work reported in~\cite{DBLP:journals/corr/abs-2012-03214} investigates combinations of star and ring architectures and proposes two forms of the TornadoAggregate algorithm: one with a ring architecture with star-based groups, the other with a star architecture with ring-based groups.
Interestingly, a substantial difference in results between the two  TornadoAggregate versions is reported. The version with star architecture and ring-based groups, outperformed the ring architecture with star-based groups. 

In~\cite{DBLP:4} D-Cliques, a topology that aims at reducing gradient bias, by grouping clients in sparsely interconnected cliques, such that the label distribution in the clique would be representative of the global distribution, is presented. This approach led to the convergence speed similar to that of a fully-connected topology with a 98\% reduction in the total number of edges, 
and 96\% reduction in the total number of messages.

A contrasting approach to data skewness mitigation, in the form of a hierarchical FL system with Federated Gradient Descent being conducted on the user-edge layer and Federated Averaging between edges and the cloud, is presented in~\cite{hierarchical}. The resulting architecture is designed with an IoT environment in mind, with the potentially less efficient connections between edge and the server supporting less frequent communication.  

Work described in~\cite{DBLP:6} uses segmentation to allow for large model training on far edge. The proposed approach relies on a combination of model segmentation level synchronization mechanisms, which divides the model into a set of not overlapping subsets, and a decentralized design reminiscent of the gossip protocol, with each worker randomly transferring the model segment to a few other workers. Model redundancy had to be included in order to ensure convergence. Discussed prototype acknowledges the problem of workers suddenly exiting and returning.
This work has been further extended in~\cite{bacombo}, forming a bandwidth-aware solution by greedily choosing a client with sufficient bandwidth to avoid delays. The convergence guarantees were provided, with the training time being reduced up to 18 times, compared to that of baselines with no accuracy degradation.

Another approach to decentralized FL (DAFCL) can be found in~\cite{DBLP:7}. In DAFCL, all clients are connected through an undirected graph. Each of them is supposed to train the model based on its local data, and exchange the results with its neighbors, through a symmetric doubly stochastic matrix. To avoid a single point of failure, the average model estimation is tracked using First Order Dynamic Average Consensus (FODAC). This architecture shows promising results. Nevertheless, to use it in Next Generation IoT environments further work on communication efficiency, and increasing resilience to sudden catastrophic events, such as user dropout, would be necessary.

In summary, research related to FL topology introduces a multitude of approaches to the problem. from the perspective of this contribution, it ``does not matter'' which topology should be used or is the best in a given scenario. The question is: how to make sure that any needed topology can be instantiated in Next Generation IoT Ecosystems. Proposing a pathway to answering this question is the goal of the remaining parts of this contribution.

\section{Federated Learning in IoT -- proposed architecture}
\label{architecture}

Let us now introduce the proposed architectural approach to Federated Learning in IoT ecosystems. Since support for different topologies has been shown to be important in large-scale real-life deployments, the possibility of easily implementing them is crucial. Moreover, the proposed architecture should be resistant to sudden user dropout, network connection with interruptions or uneven grouping of clients. 

The proposed FL architecture is developed according to the Reference Architecture (RA) introduced in the ASSIST-IoT project, and motivated by real-life scenarios, coming from three industrial pilots. This RA is based on the concept of encapsulation, in which is instantiated in the form of enablers. Interested readers should consult~\cite{assist-architecture} for necessary details. Note that the fact that the proposed FL architecture is compatible with ASSIST-IoT RA principles allows the use of additional enablers that can extend its capabilities, e.g., with a semantic toolset to enable interoperability, or self-* functionalities such as automated configuration (e.g., to control the state of topology and adjust its configuration)~\cite{10.1007/978-3-030-86359-3_32}. These aspects are, however, outside of the scope of this contribution.

\begin{figure*}[htbp]
\centering
\includegraphics[width=0.8\linewidth]{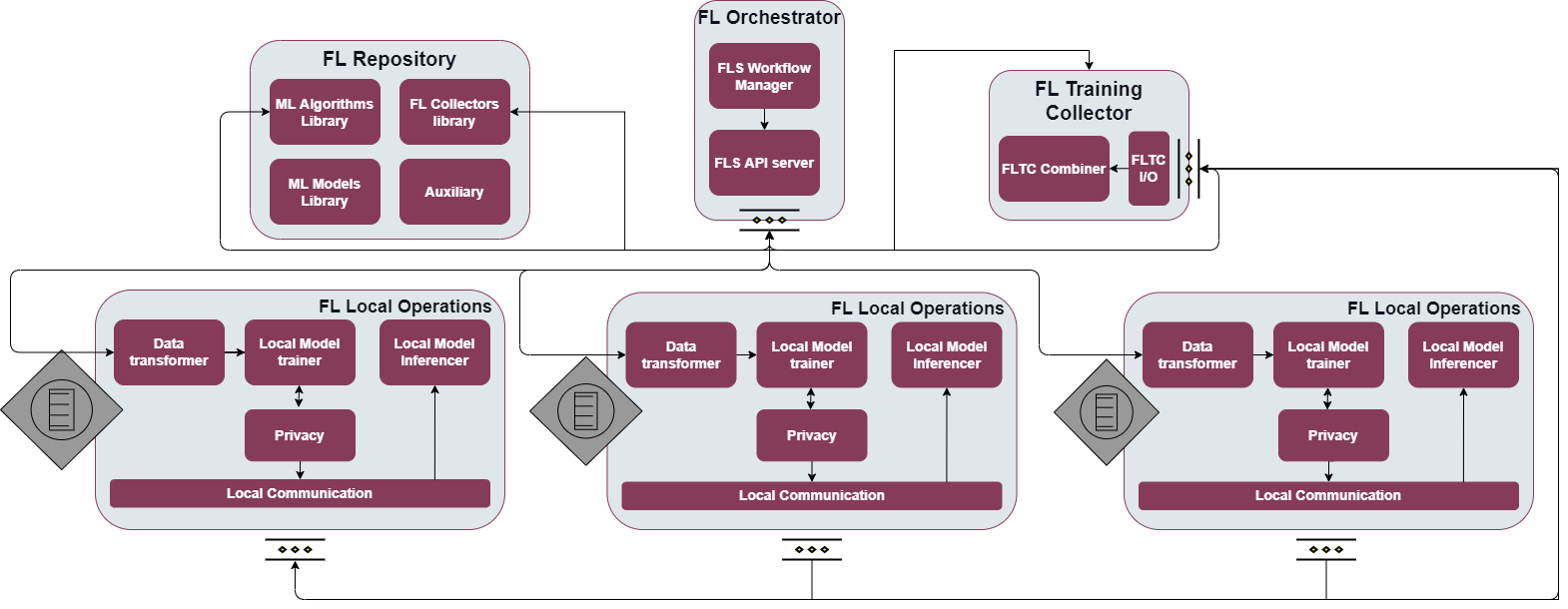}
\caption{Proposed FL in IoT architecture}
\label{FL_architecture}
\end{figure*}

As it can be seen in Fig.~\ref{FL_architecture}, the proposed FL architecture is formed by four enablers: \textit{FL Orchestrator}, \textit{FL Repository}, \textit{FL Training Collector}, and \textit{FL Local Operations}.
The \textit{FL Orchestrator} is the enabler responsible for the configuration propagation to other enablers, workflow management, and control over the FL life cycle. It also acts as the entrance gate for human interactions. Moreover, \textit{FL Orchestrator} may control the FL training process, and constraints related to e.g., the minimum number of clients, or minimum system requirements. On the other hand, the \textit{FL Repository} is a supplementary enabler for storing models, algorithms, and any data needed in the FL process. Last, the \textit{FL Training Collector} and \textit{FL Local Operations} act as FL servers and clients, respectively. They are used in the constructed system as communicating components, remaining in constant contact according to the gRPC protocol, by utilizing functionalities implemented as a part of the Flower library~\cite{flower}. In other words, the \textit{FL Training Collector} possesses the capabilities of a FL centralized server, while the \textit{FL Local Operations} (located on edge clients) has the abilities of an FL client, with the main focus placed on local model training and dataset loading. Let us now describe the \textit{FL Training Collector} and the \textit{FL Local Operations} in more detail.

\subsection{FL Training Collector}
\textit{FL Training Collector} mainly serves the role of a server node. Uploading configuration (e.g. from the \textit{FL Repository}) initiates the training process. The configuration data can include, among others, the type of aggregation algorithm used for FL, the minimal number of clients necessary in order to start training, the minimum number of clients necessary for training each round, the fraction of clients to be sampled for training or evaluation, a set timeout for the responses coming from clients, the number of clients to choose for training with blacklisting and some additional values used for later testing. The behaviour exhibited by the \textit{FL Training Collector} before and after each training, as well as evaluation round, is defined in the form of a Strategy class, in accordance with the requirements of the Flower library. This class is used by the Flower server to group clients, selected from available client interfaces, with the appropriate weights to be later sent by the server and to define the mechanisms used to aggregate results from the clients and evaluate current model performance. Due to its periodic nature (methods are called in the defined order, before and after every round), this class is also used for gathering metrics and saving current model weights, for later analysis. The metrics, which are gathered after each training round, consist of aggregated evaluation loss, global evaluation loss, and global accuracy. They are collected in order to facilitate monitoring of the training process. Later, they are locally stored in the enabler in the form of a serialized object inside a pickle file~\cite{pickle}. 

\subsection{FL Local Operations}
An instance of \textit{FL Local Operations}, the analogue for the FL client, is created similarly to the \textit{FL Training Collector}. In order to start the training, it needs to be provided with a training configuration, and the address of the \textit{FL Training Collector} instance, which it should be connected to.

\textit{FL Local Operations} enabler is responsible for loading and preprocessing the right subset of local data, and setting up the local model. It not only executes but also enhances the behaviour of an FL client in the form of classes extending the flower.client.Client class, by implementing methods of initiating, fitting the model, and evaluating the model performance. 
The evaluation accuracy and loss of the current model are computed on the local test set. The values of these metrics, in their original form, as well as an average (in the case of clustered architecture -- weighted average, in an attempt to increase the precision of the visualization, for unstable client groupings) over the metric values from all \textit{FL Local Operations} is used to assess the efficiency of the training process. Similarly to \textit{FL Training Collector}, these statistics are regularly stored as pickle files~\cite{pickle}. \textit{FL Local Operations} may also include mechanisms related to privacy, such as data encryption or differential privacy~\cite{privacy,9144394}.

\subsection{FL training process}
Let us now describe the FL training process that is to take place in the case of basic, centralized, topology.

\begin{enumerate}
    \item An instance of \textit{FL Training Collector} receives a training configuration from the \textit{FL Orchestrator}.
    \item \textit{FL Training Collector} waits for a minimal number of clients, as specified by the configuration.
    \item Required number of \textit{FL Local Operations} instances receive their training configuration, from the \textit{FL Orchestrator}, similar in content to that supplied to the \textit{FL Training Collector}, but also including identifying information about the \textit{FL Training Collector} participating in the process.
    \item Activated instances of \textit{FL Local Operations} establish a connection with the \textit{FL Training Collector}. 
    \item \textit{FL Training Collector} samples \textit{FL Local Operations} and provides them with model weights and, possibly, additional configuration, which triggers the training process
    on \textit{FL Local Operations}.
    \item \textit{FL Local Operations} instances train the model (in parallel) and return the weights along with any metrics they were requested to gather.
    \item Next, the weights are aggregated according to a strategy supplied by the \textit{FL Training Collector}. The data, along with any computed metrics, is communicated (as required) before and after model evaluation (after each round). 
\end{enumerate}   
This approach, formulated for the basic centralized architecture, can be then modified in order to support other topologies.

\section{Other FL topologies, applicability and usability}
\label{testing}

\begin{figure}[htbp]
\subfloat[Centralized architecture]{\includegraphics[width=0.9\linewidth]{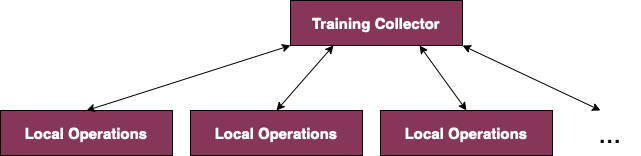}}\hfill
\subfloat[Clustered architecture]{
\includegraphics[width=\linewidth]{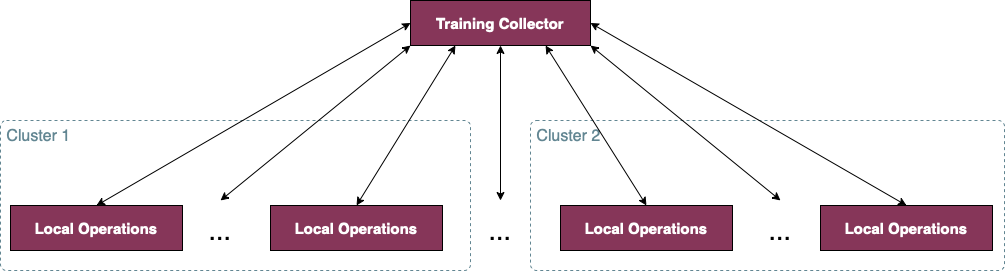}
}\hfill
\subfloat[Hierarchical architecture]{
\includegraphics[width=\linewidth]{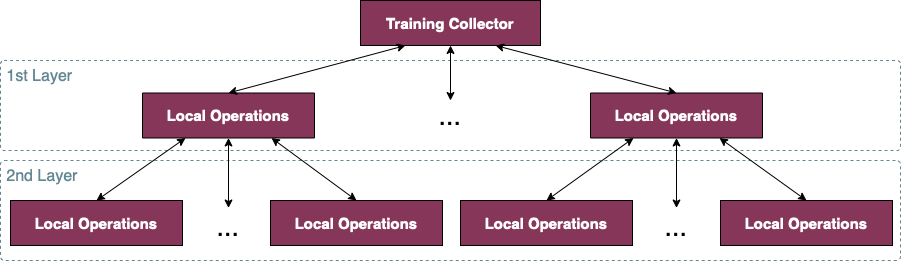}
}\hfill
\subfloat[Star architecture with ring-based groups]{
\includegraphics[width=\linewidth]{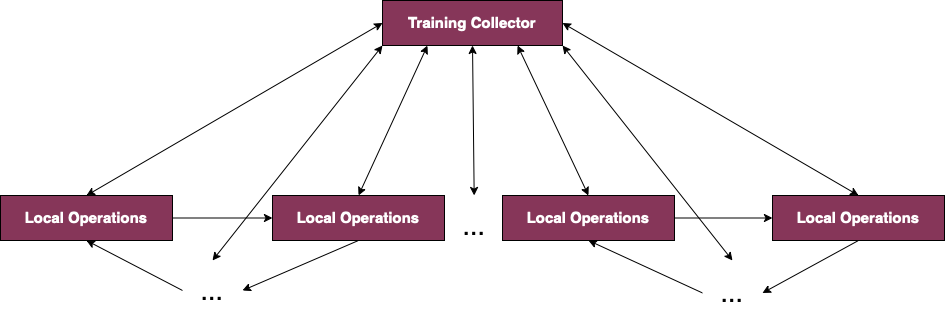}
}\hfill
\caption{ASSIST-IoT FL alternative architectures for IoT environments} 
\label{topologies_diagrams}
\end{figure}

By performing slight modifications to the basic architecture, it is possible to instantiate other topologies proposed in the literature. In particular, four topologies have been implemented and initially tried: centralized architecture, clustered architecture, hierarchical architecture, and star architecture with ring-based groups. They are illustrated in Fig.~\ref{topologies_diagrams}. It should be noted that the aim of this work was to establish that the proposed architectural approach, based on enablers originating from the ASSIST-IoT RA can be used to easily set up ``any'' FL topology. Thus, this is what was implemented and tested. The usage of these topologies for the car maintenance use case, described above, will be explored in the near future.

The basic centralized architecture was implemented following the description presented above. The possibility of using different ``parameters'' of the FL process, as represented in the setup, including client numbers, model architecture, approach to model averaging, data collected by the \textit{FL Local Operations} and \textit{FL Training Collector} has been tested. 

As for the clustered architecture, the implemented version, first, accepts a set number of clusters and then uses the Iterative Federated Clustering Algorithm (IFCA)~\cite{ifca} to dynamically determine the adherence of a given client to a cluster at the beginning of each round. Next, in the aggregation stage, the cluster models are updated, based only on the data from the clients that belong to them at the moment. When faced with IID data, the clients are determined to belong to a single cluster, which leads the architecture to behave similarly to the centralized one. For the non-IID data, the clustered architecture leads to the development of a number of models, each tailored exactly to a given cluster of clients, instead of a single global solution. The implemented architecture was tested using CIFAR-10 (for IID data) and German Traffic Sign Recognition Benchmark dataset (for non-IID data) and the results matched these found in the literature~\cite{cnn, gtrb}.

The hierarchical topology necessitates the creation of an additional component~\cite{hierarchical}. In this work it is implemented as a special case of \textit{FL Local Operations} called \textit{1st Layer Local Operations}. This necessitates that the version of \textit{FL Local Operations} acts as a basic FL client called \textit{2nd Layer Local Operations}. This additional enabler serves as FL server to \textit{2nd Layer Local Operations} and as FL client to the \textit{FL Training Collector}, aggregating the updates from the \textit{2nd Layer Local Operations} for a set number of local rounds, and afterwards propagating them to the global \textit{FL Training Collector} for aggregation. Again, the instantiated, hierarchical, topology was tested using the CIFAR-10 and German Traffic Sign Recognition Benchmark dataset and obtained results matched these reported in the literature~\cite{cnn,gtrb}.

In yet another experiment, the star topology with ring-based groups introduced decentralized elements, based on the Tornadoes architecture~\cite{DBLP:journals/corr/abs-2012-03214}. Here, the training process starts with the \textit{FL Training Collector} sending the initial model to all available \textit{FL Local Operations}. Then, the \textit{FL Local Operations} uses every local round to train the model on its local data to pass it to the next instance belonging to its ring-based group, and accept an incoming model from the previous instance, for further training. After a given number of local rounds a global aggregation (performed by the \textit{FL Training Collector} occurs. As in previous cases, the constructed topology was tested (on the CIFAR-10 and German Traffic Sign Recognition Benchmark datasets) and obtained results match these found in~\cite{cnn,gtrb}.

\begin{figure*}[htb]
\centering
\includegraphics[width=0.8\linewidth]{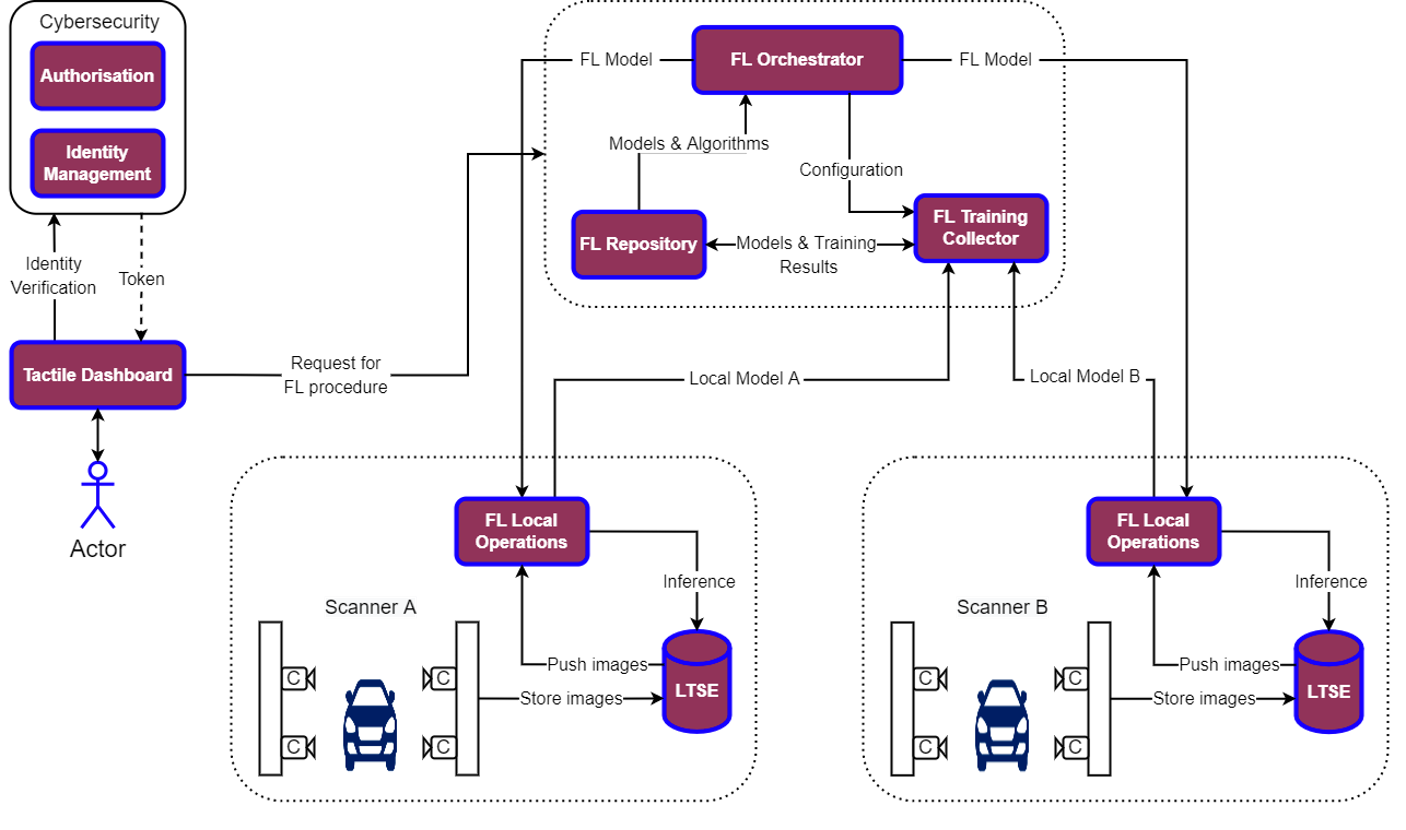}
\caption{FL architecture for the car damage use case}
\label{uc_architecture}
\end{figure*}

Finally, Fig.~\ref{uc_architecture} presents a solution envisioned for the use case described in Section~\ref{usecase} using enablers from the proposed architecture. Here, we use a centralized topology where \textit{FL Local Operations} are run on clients (cameras). 
\textit{FL Orchestrator}, \textit{FL Training Collector} and \textit{FL Repository} are located in the cloud. This environment is going to be somewhat more ``stable'', because there is a predefined number of clients in the business environment. The main goal is to distribute the processing, instead of sending all the images to the cloud and processing it centrally. Here, although the centralized topology seems to be a good choice for initial implementation, it is clear that a more complex topology will ultimately be needed.
 One of the reasons is that in extended deployment groups of scanners (one or more) may belong to different stakeholders. Therefore, a hierarchical topology would be a natural choice. Nonetheless, it has been already established (above) that such topology is easy to deliver using the existing set of enablers.

On the diagram, besides FL enablers, additional enablers designed and implemented within ASSIST-IoT (following ASSIST-IoT RA) are included addressing: cybersecurity (specifically authentication and authorization), \textit{Long Term Storage} enabler (that can provide local storage of images for FL clients), and \textit{Tactile Dashboard} (for visualizations needed in the system). These elements can provide all additional functions needed in the ecosystem.

\section{Concluding remarks}
\label{conclusions}
Even though there is a lot of research in the field of FL, most of it is devoted to FL processes, algorithms, or specific aspects such as data security. Here, we try to address issues related to the deployment of FL system in a real-life use case in an IoT ecosystem. This requires the choice of an appropriate architecture. In this context, the ASSIST-IoT RA was extended to deliver a set of enablers that allow easy configuration of FL system with machine learning parameters, as well as any required topology. Moreover, additional enablers, created for the RA allow turning the FL process into a complete, robust solution.

\bibliographystyle{IEEEtran}
\bibliography{conference_101719}

\end{document}